\documentclass[10pt,twocolumn]{article}

\usepackage{iccv}
\usepackage{times}
\usepackage{epsfig}
\usepackage{graphicx}
\usepackage{amsmath}
\usepackage{amssymb}
\usepackage[english]{babel}
\usepackage[utf8]{inputenc}
\usepackage{xcolor}
\usepackage{multicol}
\usepackage[ruled,vlined]{algorithm2e}
\usepackage{caption}
\usepackage{tabularx}
\usepackage{bbm}
\usepackage{capt-of,etoolbox}
\usepackage{microtype}
\usepackage{enumitem}
\usepackage{booktabs} 
\usepackage[toc,page]{appendix}

\usepackage[pagebackref=true,breaklinks=true,colorlinks,bookmarks=false]{hyperref}

\iccvfinalcopy 

\usepackage{geometry}
\usepackage[accsupp]{axessibility} 
\ificcvfinal\fi

\begin{document}

\title{PINs: Progressive Implicit Networks for Multi-Scale Neural Representations}

\author{Zoe Landgraf$^{1}$, Alexander Sorkine Hornung$^{2}$, Ricardo Silveira Cabral$^{2}$
\thanks{The research presented in this paper was done during an internship at Meta, Switzerland}
\thanks{$^{1}$Department of Computing, Imperial College London, UK. \tt zoe.landgraf15@imperial.ac.uk}
\thanks{$^{2}$ Meta, Switzerland}
}

\maketitle
\ificcvfinal\thispagestyle{empty}\fi

\begin{abstract}
Multi-layer perceptrons (MLP) have proven to be effective scene encoders when combined with higher-dimensional projections of the input, commonly referred to as \textit{positional encoding}. 
However, scenes with a wide frequency spectrum remain a challenge: choosing high frequencies for positional encoding introduces noise in low structure areas, while low frequencies result in poor fitting of detailed regions. To address this, we propose a progressive positional encoding, exposing a hierarchical MLP structure to incremental sets of frequency encodings.
Our model accurately reconstructs scenes with wide frequency bands and learns a scene representation at progressive level of detail \textit{without explicit per-level supervision}. The architecture is modular: each level encodes a continuous implicit representation that can be leveraged separately for its respective resolution, meaning a smaller network for coarser reconstructions. 
Experiments on several 2D and 3D datasets show improvements in reconstruction accuracy, representational capacity and training speed compared to baselines. 
\end{abstract}

\vspace{-5mm}
\section{Introduction}
\label{sec:introduction}

\begin{figure}[ht!]
    \centering
    \begin{minipage}[c]{0.99\linewidth}
    \includegraphics[width=\linewidth]{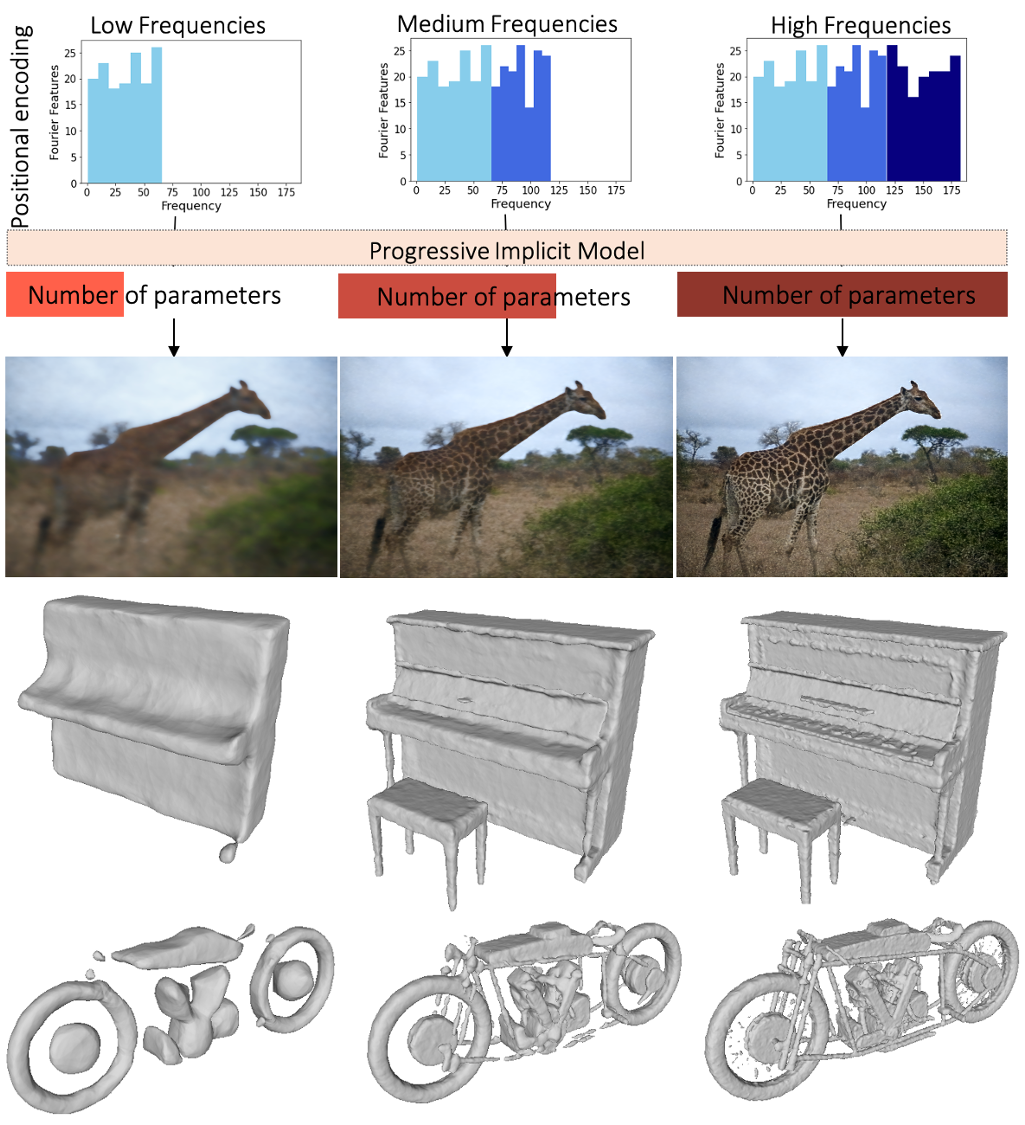}
    \end{minipage}
    \vspace{-1.5mm}
    \caption{\small{Method overview. The input coordinates are projected by incremental sets of frequency encodings (\textbf{top row}) and processed by a hierarchical MLP structure which produces reconstructions in 2D (\textbf{row $2$}) and 3D (\textbf{row $3$ and $4$}) at progressive level of detail.}}
    \vspace{-5mm}
    \label{fig:teaser_image}
\end{figure}

\begin{figure*}[ht!]
    \centering
    \begin{minipage}[c]{1.0\linewidth}
    \includegraphics[width=\linewidth]{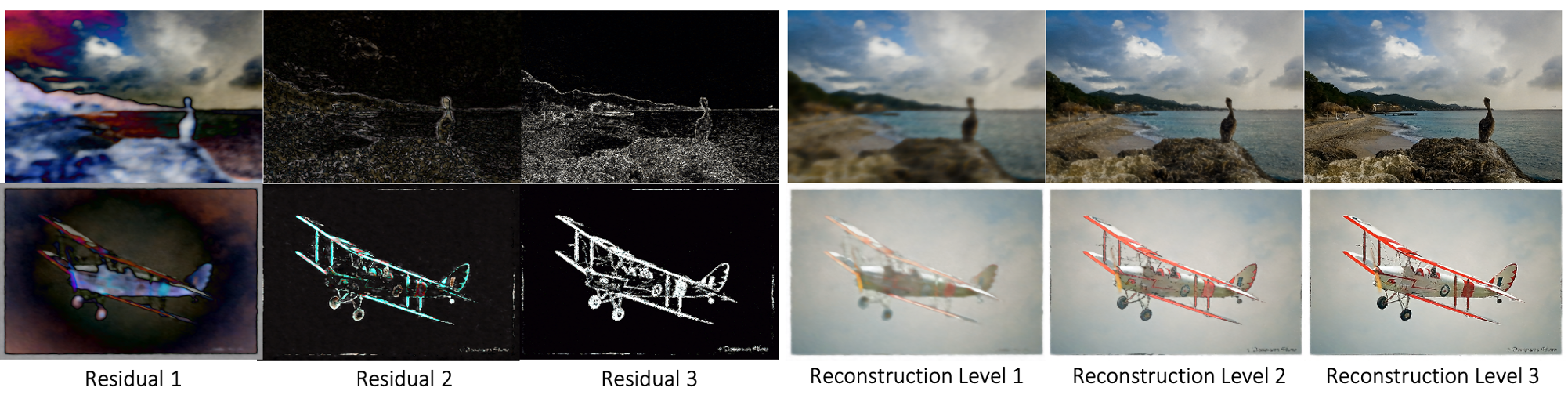}
    \end{minipage}
    \vspace{-1mm}
    \caption{\small{2D reconstruction example. \textbf{Left:} Image residuals predicted by the network \textbf{Right:} Progressive reconstruction levels obtained from combining the base image $c$ and the respective image residuals}}
    \vspace{-3mm}
    \label{fig:reconstruction_example_2D}
    
\end{figure*}

Neural Implicit Functions are gaining popularity as alternative 2D image and 3D shape representations. Using a simple MLP encoder, these networks approximate a function mapping between spatial coordinates and a quantity of interest such as colour, occupancy or SDF values. They have proven to be very effective at fitting natural images \cite{tancik2020fourfeat,Martel2021ACORNAC} and 3D shapes \cite{Mescheder2019OccupancyNL, Park2019DeepSDFLC, Chabra2020DeepLS} and have been applied to various computer vision tasks, including novel view synthesis \cite{Niemeyer2020DifferentiableVR, Mildenhall2020NeRFRS} and generative shape modelling \cite{Chen2019LearningIF}.

However, as simple MLP encoders, neural implicit functions suffer from spectral bias \cite{Rahaman2019OnTS, Basri2020FrequencyBI}, which prevents them from 
learning high frequency detail in signals. Projecting the input onto a manifold containing high frequency components however, reduces the spectral bias \cite{Rahaman2019OnTS}. Recent works \cite{Mildenhall2020NeRFRS, Zhong2020ReconstructingCD} have demonstrated this experimentally, mapping the input through a set of sinusoidal functions (\textit{positional encodings}). In a parallel approach, \cite{sitzmann2019siren} use periodic activation functions instead of ReLUs to enable MLPs to learn high frequency content. \cite{tancik2020fourfeat} propose an improved positional encoding based on Fourier Features \cite{Rahimi2007RandomFF}
and show that 
one can essentially \textit{tune} the range of frequencies that can be learnt by an MLP through the frequencies in the positional encoding. However, MLPs with these encodings struggle to fit signals with a wide frequency spectrum: high frequencies in the encoding enable fitting small detail in a signal, but introduce noise in smoother regions. To improve on that, \cite{hertz2021sape} propose a training scheme based on a spatial mask, which gradually introduces frequency encodings to the network. Whilst achieving compelling results, this method requires spatial masks, which don't scale well and reduce training speed, and its training scheme for unwrapping frequencies is non-differentiable and requires manual hyperparameter tuning.

We show that a continuous function can learn to reconstruct signals with wide frequency ranges and in a progressive fashion (see Figure \ref{fig:teaser_image}): 
we partition the representation into hierarchical levels, each receiving a subset of our positional encoding, whereby lower layers process lower frequency sets compared to higher layers. Our final reconstruction is a simple composition of all intermediate levels. This results in a model that is trainable end-to-end and we find that our architecture naturally induces lower layers to learn a coarser reconstruction while higher layers focus on adding details to the scene (see Figure \ref{fig:reconstruction_example_2D}), without explicit per-level supervision.
Our method provides progressive level of detail while yielding on par or superior reconstruction quality to baselines. In addition, our architecture is modular - for a coarser representation, one can drop the MLPs of higher levels and reconstruct the scene with only a small portion of the parameters. Overall our method provides:
\vspace{-1mm}
\begin{itemize}[leftmargin=*]
    \item A multi-scale representation based on hierarchical implicit functions with progressive positional encoding, providing incremental level of detail.
    \item Improved reconstruction quality, in particular for scenes with a wide frequency spectrum.
    \item End-to-end trainable model without per-level supervision
\end{itemize}
\vspace{-3mm}

\begin{figure*}[ht!]
    \centering
    \begin{minipage}[c]{0.999\linewidth}
    \includegraphics[width=\linewidth]{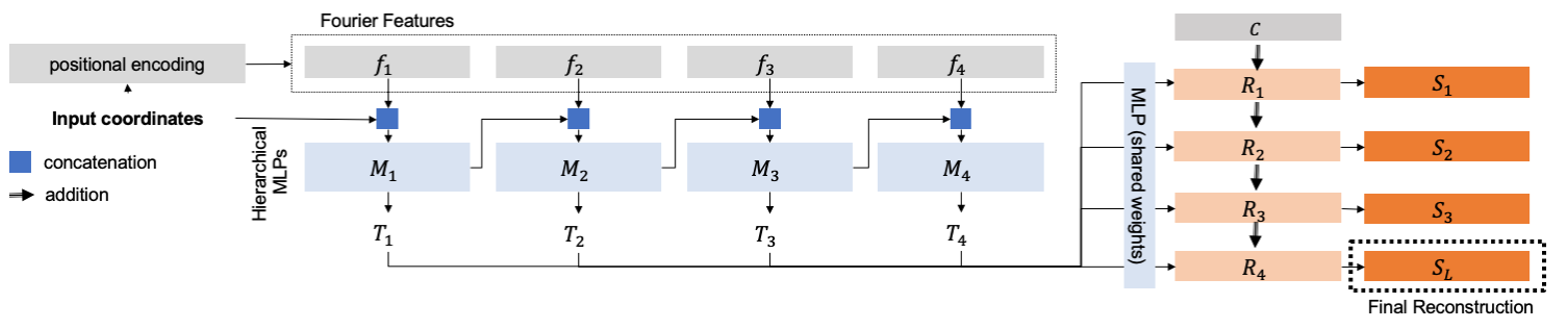}
    \end{minipage}
    \vspace{-7mm}
    \caption{\small{Model architecture for $4$ levels of detail.}}
    \vspace{-2mm}
    \label{fig:architecture}
\end{figure*}

\section{Related Work}
\label{sec:related_work}

\paragraph{Neural implicit representations}
First introduced as novel shape representations for occupancy \cite{Mescheder2019OccupancyNL} and SDFs  \cite{Park2019DeepSDFLC}, Neural Implicit Functions have become a popular alternative to classical 3D representations. Several works have used them for single- view shape inference  \cite{pifuSHNMKL19, Liu2019LearningTI}, shape decomposition \cite{Deng2020CvxNetLC} and instance-aware SLAM systems \cite{Li2020FroDOFD}. They have been shown to support larger scene representations \cite{Chabra2020DeepLS, Huang2020DIFusionOI}, when composed as a voxelgrid of small implicit functions representing local detail. \cite{Genova2020LocalDI} demonstrate that compositional implicit functions also yield improved reconstruction accuracy for smaller shapes. Recent work has explored their use for image synthesis as object-specific implicit fields \cite{Niemeyer2021GIRAFFERS}, as well as for dynamic scene graphs \cite{Ost2021NeuralSG}. 
Combined with differentiable volumetric rendering \cite{DVR}, neural implicit representations gave rise to an explosion of novel view synthesis approaches \cite{Mildenhall2020NeRFRS}. Other works address generalisation, conditioning the neural implicit representation on features extracted from images \cite{Yu2021pixelNeRFNR} or on latents that encode shape priors \cite{Schwarz2020NEURIPS, jang2021codenerf}. A few approaches have focused on improving training and rendering speed \cite{Reiser2021KiloNeRFSU, yu2021plenoctrees} and the representation itself: to address the spectral bias of neural implicit representations based on a simple MLP, periodic activation functions \cite{sitzmann2019siren} and positional encodings \cite{Rahaman2019OnTS, tancik2020fourfeat, hertz2021sape} were proposed.
Similarly to Park \etal \cite{Park2019DeepSDFLC}, our method is designed as a scene representation, in 2D and 3D. Instead of using a single MLP for scene representation we propose a hierarchical structure of small MLPs composing the scene at progressive level of detail. Similarly to Hertz \etal \cite{hertz2021sape}, we use Fourier Features as positional encodings, but instead of a spatial frequency mask, we condition each of the MLPs in our hierarchical structure on subsets of these encodings. 
\vspace{-2.5mm}
\paragraph{Multi-scale neural implicit representations}
A few recent approaches propose to add hierarchical or multi-scale structure to neural implicit representations; they can be divided into those, which use a form of space partitioning and those, that use a frequency based approach. Of the space partitioning methods, most approaches leverage Octrees and have as primary goal a more efficient rendering or training time \cite{yu2021plenoctrees, Martel2021ACORNAC, tang2021octfield}.
The method presented by Chen \etal \cite{chen2021multiresolution} generates a multi-scale representation: A feature encoder generates a hierarchy of latent code grids, which represent the scene at different resolutions in a lower dimensional manifold. To reconstruct the scene at a specific resolution, the latent grid is interpolated based on continuous coordinate samples and decoded using a neural implicit function. The methods based on frequency-decomposition add multi-scale structure in \textit{continuous function space}: Wang \etal \cite{Wang2021GeometryConsistentNS} use a composition of two SIREN models \cite{sitzmann2019siren}, to represent 3D shapes as implicit displacement fields: a smooth base surface is refined with predicted displacements along the surface normal of the base surface. 
Lindell \etal \cite{lindell2021bacon} propose a multi-scale representation based on multiplicative filter networks (MFN) \cite{fathony2021multiplicative}, a simple linear combination of Fourier or Gabor wavelet functions applied to the input. To reconstruct at multiple resolutions, linear layers are added at different depths of the MFN, to generate intermediate outputs. These outputs, when supervised, generate reconstructions at increasing level of detail.
Similarly to Wang \etal \cite{Wang2021GeometryConsistentNS} and Lindell \etal \cite{lindell2021bacon}, we model our multi-scale representation in continuous function space. However, our architecture is based on MLPs with ReLU activations, allows for multiple levels of detail and we constrain the representation of each level through frequencies in a Fourier Feature encoding of the input. Contrary to Lindell \etal \cite{lindell2021bacon}, our intermediate levels compose the final reconstruction and do not require supervision during training.

\section{Preliminaries}
\label{sec:preliminaries}

\textbf{Neural Implicit Representations}
encode signals in continuous space using a neural network parametrisation. Often referred to as coordinate-based networks, they are defined as a function mapping $f: \textbf{x} \rightarrow V$, with $\textbf{x} \in {\rm I\!R}^{1,2,3}$; $V$ is a quantity of interest such as colour or occupancy. In 3D, the surface is modelled as the levelset of a continuous function $f:f(\textbf{x}) = 0$. By design, the representation predicts a single point in the signal's domain at a time and multiple values can be obtained by querying the representation at the corresponding set of coordinates $\{\textbf{x}_{1} \dots \textbf{x}_{N}\}$. 

\textbf{Positional Encoding} was first introduced in Natural Language Processing \cite{Vaswani2017AttentionIA} to inject information about the relative position of word tokens into the Transformer Model. In the context of Neural Implicit Representations, positional encoding refers to the projection of the input (spatial coordinates) to a higher dimensional space $P: {\rm I\!R}^{1, 2, 3} \rightarrow {\rm I\!R}^{N}$ \cite{Mildenhall2020NeRFRS}. Several function mappings have been proposed, including simple sinusoidal mappings to fit neural radiance fields \cite{Mildenhall2020NeRFRS} and reconstructing 3D protein structures \cite{Zhong2020ReconstructingCD}, non-axis aligned Fourier basis functions: $'Fourier Features'$ \cite{tancik2020fourfeat} and a positional encoding based on multi-scale B-Splines \cite{Wang2021SplinePE}.
\textbf{Fourier Features}
 have become a popular framework for fast kernel method approximations \cite{Li2019TowardsAU}. First proposed by Rahimi \etal \cite{Rahimi2007RandomFF}, the approximation is based on Bochner's theorem \cite{Bochner}, which shows that any continuous shift-invariant kernel is the Fourier Transform of a positive bounded measure (\textit{spectral measure}). The approach proposed by Rahimi \etal \cite{Rahimi2007RandomFF} approximates such a continuous kernel by its Monte-Carlo estimate using samples from this spectral measure: \textit{Fourier Features} \cite{Li2019TowardsAU}. Tancik \etal \cite{tancik2020fourfeat} relate this method to the frequency mapping applied to inputs of neural implicit functions and propose Fourier Features as an improved form of positional encoding: a set of Fourier bases $\textbf{F} : \{cos(\omega_{i} x + b_{i}) \dots \ cos(\omega_{n} x + b_{n}\}$ where $\omega$ and $b$ are sampled from a parametric distribution (e.g $ \mathcal{N}(\mu=0,\,\sigma)$). As a positional encoding, Fourier Features map input coordinates to a higher dimensional manifold as $P: \textbf{x} \in {\rm I\!R}^{1,2,3} \rightarrow \textbf{F(x)} \in  {\rm I\!R}^{n}$.
We choose a Fourier Feature mapping as positional encoding for our progressive implicit representation and refer to it in the rest of this paper as Fourier Features (FF).

Tancik \etal \cite{tancik2020fourfeat} show that the frequencies of the input mapping function directly affect the spectral falloff of the Neural Tangent Kernel: positional encoding influences how fast, and if, specific frequencies of a signal can be learnt. They show through parameter search that sampling the frequencies of their proposed Fourier Feature encoding at $\sigma = 10$ generates the best reconstruction of natural images; however, no connection is made to the frequency composition of the target signal. We perform a similar parameter search for $\sigma$ (see Section \ref{sec:sigma_sensitivity}) and provide a short analysis relating $\sigma$ to the frequency composition of the target signal in Appendix \ref{app:FF_FT}.
\begin{figure*}[ht!]
    \centering
    \begin{minipage}[c]{0.99\linewidth}
    \includegraphics[width=\linewidth]{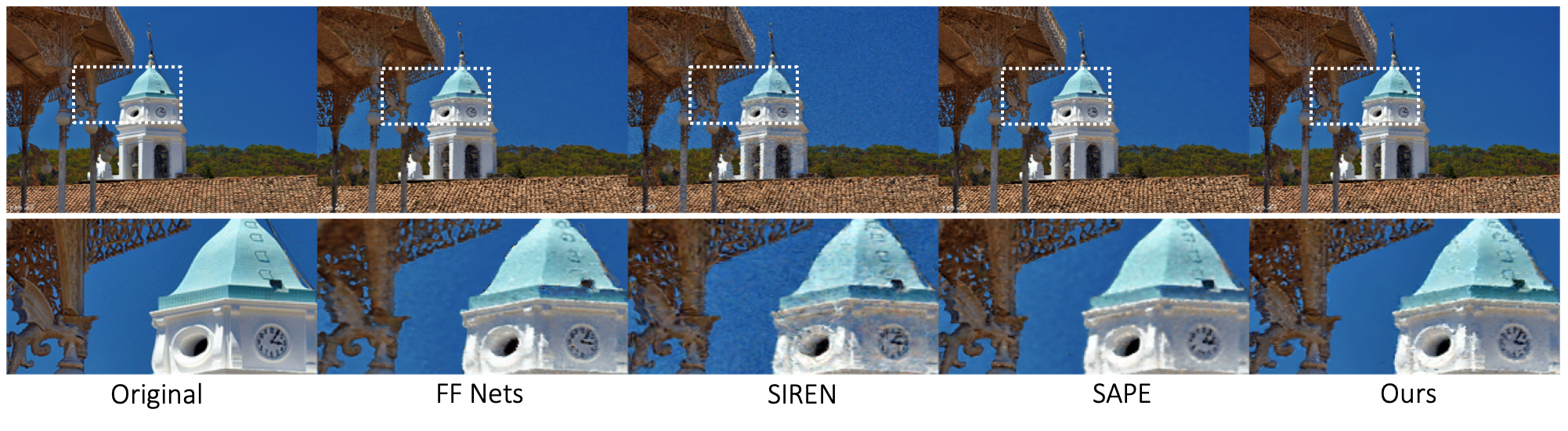}
    \end{minipage}
    \vspace{-4mm}
    \caption{\small{Qualitative results on a 2D image regression task. We compare our model against baselines on images from the COCO $2017$ validation set \cite{Lin2014MicrosoftCC}}
    }
    \vspace{-4mm}
    \label{fig:2D_regression_qualitative}
\end{figure*}
\vspace{-1mm}
\section{Method}
\label{sec:method}
\subsection{Progressive Fourier Feature encoding}
We formulate the task of reconstructing a scene $S$ as a composition of a base component $c$ and a set of residuals $R_{1 \dots L}$: 
\begin{equation}
\label{eq:scene_composition}
    S(\textbf{x}) = c + \sum_{l=1}^{L}{\omega_{l} R_{l}(f_{l}(\textbf{x})),}
\end{equation}
where $\textbf{x}$ is a coordinate vector and each residual $R_{l}$ is parameterised by an MLP. The contribution of $R_{l}$ to the final representation is controlled by $w_{l}$, and $c$ is a constant set according to the reconstruction task domain. $R_{l}$ takes as input a subset $f_{l}(\textbf{x})$ of Fourier Feature mapping $\textbf{F}$ with a set $\textit{F}$ of frequencies sampled from a Gaussian distribution $\textit{F} \sim \mathcal{N}(0, \sigma)$. The sampled frequencies are sorted by increasing value and divided into $L$ subsets $\{f_{1 \dots L}\}$. By definition, each subset $f_{l}$ now contains frequencies $\{k_{1 \dots N}\}$ which are sorted by increasing value and for any two adjacent subsets, $f_{l}$ and $f_{l-1}$, $k_{l,1} > k_{l-1,N}$, whereby $k_{l,1}$ and $k_{l-1,N}$ are the first and last frequency of the subsets $l$ and $l-1$ respectively.
Intermediate levels of reconstruction $S_{l^{'}}$ can be obtained by composing part of the residuals: $S_{l^{'}}(\textbf{x}) = \sum_{l=0}^{l^{'}}{\omega_{l} R_{l}(f_{l}(\textbf{x}))}$. Intuitively, the residual $R_{l}$ learnt by one specific level is guided by the frequency encodings it is exposed to. Given the structure of our progressive Fourier Feature encoding $\{f_{1 \dots L}\}$, the first residuals are encouraged to focus on low frequency content while later residuals focus on high frequency content in the scene. 

Inspired by the fact that power spectral densities of natural signals decay exponentially (i.e., larger proportions of the signal come from lower frequency ranges) \cite{esp}, we reduce the weight of higher levels compared to lower levels for the final reconstruction.
Empirically, we set $w$ to a decreasing geometric progression such that 
$w_{l} = \frac{1}{l + 2}$ 
and show that this leads to better results than equal weightings of $R_{l}$ in our ablation studies.

In 1D, our model maps $x$ to $y$ coordinates: $S: x \rightarrow y$ and we set $c = 0$. In 2D, coordinates are mapped to pixel values: $S: x,y \rightarrow P(x,y)$ and we set $c$ to the image mean $\overline{m}$ (detailed explanation provided in section \ref{sec:2D_image_regression}). In 3D, our model defines a levelset as $V(\tau) = \{x: S(x) = \tau\}$
where $V$ is the volume containing the 3D shape whose surface lies at $V(0)$, $x$ is a coordinate in $V$ and $S$ maps coordinates to SDF values $\tau$ as $S: {\rm I\!R}^{3} \rightarrow {\rm I\!R}$.
Similarly to \cite{chen2021multiresolution}, our residuals $R_{l}$ are defined as $R_{l} = S_{l} - S_{l-1}$. However, in our formulation, we set $S_{0}$ to be a constant $c = 0$.
\subsection{Architecture}
Our architecture is composed of multiple stacked small MLPs $M_{1 \dots L}$, whereby each MLP receives as input a set of FF mappings $f_{l}(\textbf{x})$ and the output of the previous level (level-conditioning). The first level receives as input $f_{1}(\textbf{x})$ and the raw coordinates $\textbf{x}$. The output of each level is a feature tensor $T_{l}$:
\begin{equation}
\label{eq:architecture}
\begin{split}
    M_{l}: f_{l}(\textbf{x}), T_{l-1} \rightarrow T_{l}, \quad l > 0 \\
    M_{l}: f_{l}(\textbf{x}), \textbf{x} \rightarrow T_{l}, \quad l = 0
\end{split}
\end{equation}
Each feature tensor $T_{l}$ is then mapped to a residual $R_{l}$ by another MLP which shares weights across each level and acts as a feature to domain mapping. Finally, the network outputs are composed into the final reconstruction according to Equation \ref{eq:scene_composition}. Using an intermediate feature representation $T_{l}$ at each level allows for a more expressive feature representation to be passed to the next level during level-conditioning. We experimented with per-level domain mapping but found no explicit benefit over using a single MLP with shared weights. 
While the number of layers $L$ as well as the number and range of frequencies $\textit{F}$ are selected before training,
we only apply a loss on the final reconstruction $S$, leaving it up to the network how to decompose the scene representation into progressive levels: the decomposition into levels of detail is therefore \textit{unsupervised}.
Intuitively, our architecture motivates a decomposition into progressive levels of detail by restricting the frequency range each level-specific MLP has access to. We show in our ablation studies that the conditioning of $M_{l}$ on the previous level $M_{l-1}$ yields higher reconstruction accuracy. \textbf{Per-level modularity} The design of our architecture allows for each level of reconstruction to be used independently at test time. For a reconstruction at an intermediate level of detail $S_{l=2}$, MLPs $M_{3 \dots L}$ can be dropped and the reconstruction will be computed as $S_{l=2}(\textbf{x}) = \sum_{l=0}^{l=2}{\omega_{l} R_{l}(f_{l}(\textbf{x}))}$.   An overview of our method and the architecture are provided in Figures \ref{fig:teaser_image} and \ref{fig:architecture}.
\subsection{Loss}
To train our model, we apply a reconstruction loss $L_{r}$ at the final reconstruction $S$, as well as a regularisation loss $L_{reg}$ which encourages intermediate levels of detail $S_{l^{'}}$ to be close to the  ground truth scene $S_{g}$:
\vspace{-3mm}
\begin{equation}
    L_{reg} = \sum_{l=1}^{N}{L_{r}(S_{L}(\textbf{x}),S_{g}).}
\end{equation}
\vspace{-0.2mm}
Our final loss is defined as $L_{r} + \omega L_{reg}$ and we find that a value of $\omega = 0.01$ works well for our experiments. We experiment with simple $L1$ and $L2$ norms for $L_{r}$, as well as a perceptual loss based on VGG features. Perceptual losses based on deep network features have to be regularised using the $L1$ or $L2$ norm \cite{Johnson2016PerceptualLF} and we find that although the perceptual loss yields slightly more pleasing visual results for 2D regression tasks, the difference is not significant enough to motivate the introduction of additional hyperparameters. For our final experiments we therefore use the $L2$ norm.
\vspace{-0.1mm}
\subsection{Training and implementation details}
We use the Adam optimiser \cite{Kingma2015AdamAM} and a standard learning rate of $1\mathrm{e}{-3}$ for all experiments. Unless mentioned otherwise, we train on a uniformly sampled subset of $50\%$ of the image pixels for all 2D regression tasks. For 3D shape regression, we train on an average of $456k$ SDF samples per shape. For the presented experiments, unless specified, we train with $3$ levels of detail, a hidden layer size of $256$ per level and $\sigma=15$. We extend the baseline architectures to have the same number and size of hidden layers.
\begin{figure*}[ht!]
    \centering
    \begin{minipage}[c]{0.95\linewidth}
    \includegraphics[width=\linewidth]{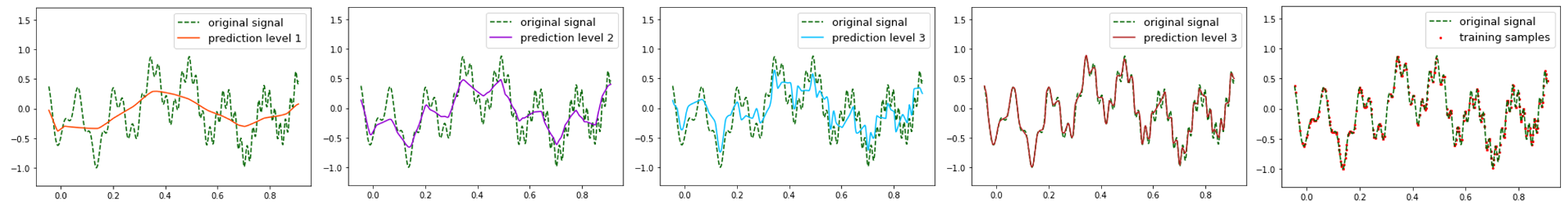}
    \end{minipage}
    \vspace{-1.5mm}
    \caption{\small{Fitting 1D periodic signals. \textbf{Left} to \textbf{right}: Reconstruction Levels 1-4 and original signal with training samples.}}
    \label{fig:1D_experiments} 
    \vspace{-2.5mm}
\end{figure*}
\begin{figure*}[ht!]
    \centering
    \begin{minipage}[c]{0.99\linewidth}
    \includegraphics[width=\linewidth]{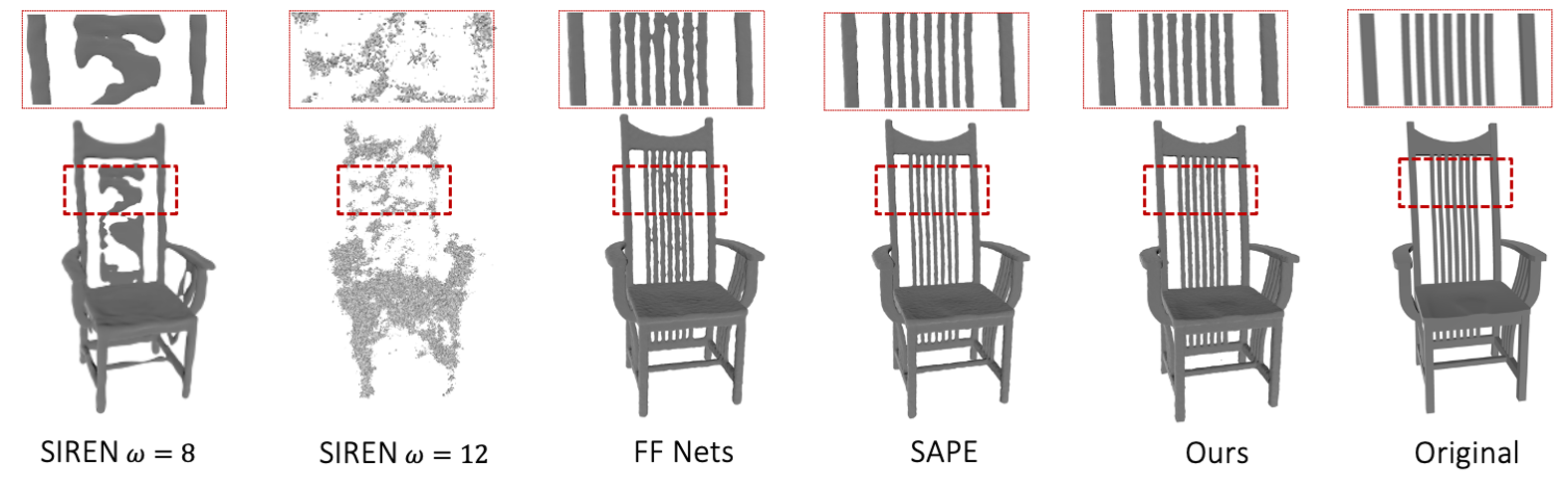}
    \end{minipage}
    \vspace{-3.5mm}
    \caption{\small{Qualitative results for a 3D regression task. We regress a TSDF and compare against SIREN models with two different $\omega$ and set $\sigma$ to $2$ for our model and the other baselines using Fourier Features.}}
    \vspace{-2mm}
    \label{fig:3D_regression_results_chair}
\end{figure*}
\section{Experiments}
We evaluate our method with several 1D, 2D and 3D regression tasks as well as in terms of representational capacity and training time. We also provide a set of ablation studies and an analysis on the relation between positional encoding frequencies and signal frequencies.
\textbf{Baselines}
We compare our method against 3 different baselines: 1) A simple MLP with Fourier Feature encoding (FF Net) \cite{tancik2020fourfeat} 2) SIREN, an MLP with Sinusodial activations \cite{sitzmann2019siren} 3) SAPE, an MLP with spatially-adaptive progressive encoding with a mask resolution of 64 \cite{hertz2021sape}. For SIREN, we set $w=30$ as suggested by Sitzmann \etal \cite{sitzmann2019siren}. We observe that for regressing 3D shapes, $w=30$ leads to divergence and choose lower values between $5$ and $10$. We also observe SIREN to be unstable if optimising for too many iterations and reduce the number of iterations to achieve the best possible reconstruction.
\subsection{1D signal regression}
In a first instance, we qualitatively evaluate our model on periodic signals composed of multiple sinusoids. As can be seen from the example in Figure \ref{fig:1D_experiments}, 1D signals are fit at progressive level of detail by our model. Examples of the target signals and more results are provided in Appendix \ref{app:1D_2D_qual_results}.
\begin{table*}[ht!]
\begin{center}
\resizebox{1.0\textwidth}{!}{%
\begin{tabular}{ l c c c c c c c }
 \toprule
 Model & lamp (ChD $\downarrow$) & car (ChD $\downarrow$) & chair (ChD $\downarrow$) & sofa (ChD $\downarrow$) & motorbike (ChD $\downarrow$) & bed  (ChD $\downarrow$) & camera (ChD $\downarrow$) \\
 \midrule
 FF Nets & 2.5$\pm3.4\mathrm{e}{-3}$ & 2.1$\pm4.8\mathrm{e}{-4}$& 0.92$\pm 1.5\mathrm{e}{-4}$ & 1.5$\pm5.6\mathrm{e}{-4}$& \textbf{1.0$\pm$1.5$\mathrm{e}{-5}$} & 2.83$\pm5.2\mathrm{e}{-3}$ & 3.46$\pm1.26\mathrm{e}{-2}$  \\ 
 SIREN & 25.4$\pm4.2\mathrm{e}{-3}$ & 2.2$\pm 5.2\mathrm{e}{-4}$ & 28.4$\pm 6.7\mathrm{e}{-3}$ & 1.6$\pm5.5\mathrm{e}{-4}$ &  1.7$\pm6.4\mathrm{e}{-4}$ & 3.22$\pm6.9\mathrm{e}{-3}$ & 2.31$\pm2.68\mathrm{e}{-3}$  \\ 
 SAPE & 6.7$\pm9.2\mathrm{e}{-2}$ & 2.2$\pm6.6\mathrm{e}{-4}$& 1.50$\pm1.1\mathrm{e}{-3}$ & 6.6$\pm$0.42 & 2.8$\pm$3.9$\mathrm{e}{-4}$ & 4.58$\pm2.4\mathrm{e}{-2}$ & 4.38$\pm2.7\mathrm{e}{-2}$ \\ 
 Ours & \textbf{1.5$\pm$1.0$\mathrm{e}{-4}$} & \textbf{2.0$\pm$5.1$\mathrm{e}{-4}$} & \textbf{0.87$\pm$1.7$\mathrm{e}{-4}$} & \textbf{1.48$\pm$5.4$\mathrm{e}{-4}$} & 1.1$\pm $6.7$\mathrm{e}{-5}$ & \textbf{2.79$\pm$4.7$\mathrm{e}{-3}$} & \textbf{2.05$\pm$2.7$\mathrm{e}{-3}$} \\ 
 \bottomrule
\end{tabular}
}
\end{center}
\vspace{-0.5cm} 
\caption{\small{Evaluation on 3D models from 3D Warehouse \cite{3DW} in terms of the bi-directional Chamfer Distance ($mm$)}}
\vspace{-1mm}
\label{tab:shape_regression}
\end{table*}
\subsection{2D image regression}
\label{sec:2D_image_regression}
\begin{figure*}[ht!]
    \centering
    \begin{minipage}[c]{0.99\linewidth}
    \includegraphics[width=\linewidth]{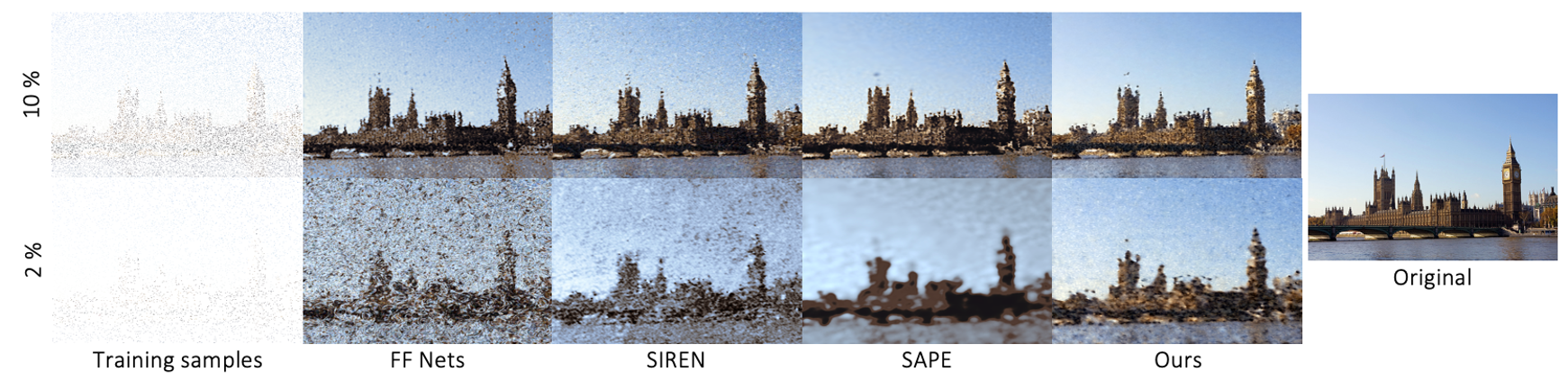}
    \end{minipage}
    \vspace{-3.5mm}
    \caption{\small{Fitting natural images with sparse sample sets of $2\%$ (\textbf{top row}) and $10\%$ (\textbf{bottom row}) of all image pixels.}}
    \label{fig:sample_sparsity} 
    \vspace{-3.5mm}
\end{figure*}

We evaluate our method on natural image reconstruction tasks with a subset of the ImageNet test dataset \cite{deng2009imagenet}, high resolution images from the DIV2K validation dataset \cite{Agustsson2017ChallengeOS} and qualitatively, with images from the $2017$ COCO validation set \cite{Lin2014MicrosoftCC}.
We evaluate in terms of Mean-Squared-Error (MSE), Peak Signal to Noise Ratio (PSNR) as well as a perceptual loss based on the L2 norm between VGG features \cite{Johnson2016PerceptualLF}. Our quantitative results (see Table \ref{tab:image_regression}) show that we outperform SIREN and the FF Net baseline in all metrics; SAPE outperforms our method on the perceptual loss for the ImageNet dataset. Qualitative results can be found in Figures
\ref{fig:teaser_image}, \ref{fig:2D_regression_qualitative}, \ref{fig:reconstruction_example_2D} and Appendix \ref{app:1D_2D_qual_results}. Our method reconstructs images with less noise, particularly for scenes with wide frequency bands. Thanks to the continuous representation of every level of detail, we are able to handle very fine detail against plain backgrounds (see Figure \ref{fig:2D_regression_qualitative}). Although this would in theory be possible with SAPE with a sufficiently high frequency mask resolution ($>64$), the training time would be too slow for practical applications (see results of section \ref{sec:model_size_and_training_time}). 
\begin{figure*}[ht!]
    \centering
    \begin{minipage}[c]{0.99\linewidth}
    \includegraphics[width=\linewidth]{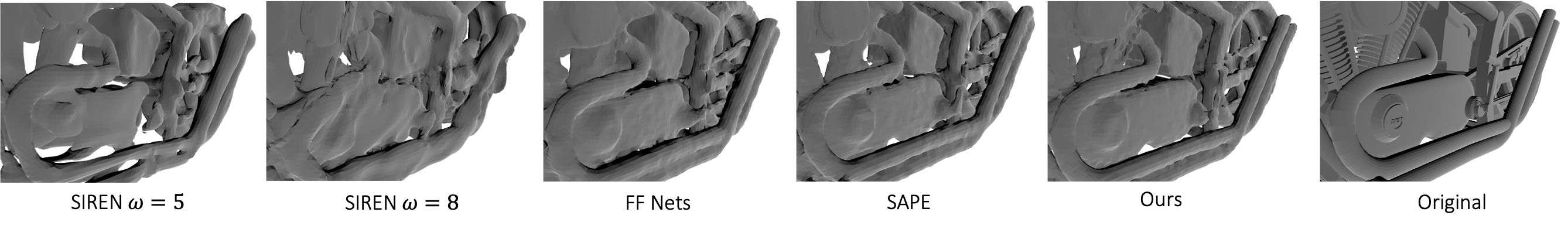}
    \end{minipage}
    \vspace{-2.5mm}
    \caption{\small{Qualitative results for a 3D regression task. We regress a TSDF and compare against SIREN models with two different $\sigma$ and set $\sigma$ to $2$ for our model and the baselines using Fourier Features.}}
    \vspace{-2mm}
    \label{fig:3D_regression_results_motorcycle}
\end{figure*}
\begin{table}
\begin{center}
\resizebox{0.48\textwidth}{!}{%
\begin{tabular}{l c c c }
 \toprule
\multicolumn{1}{l}{Model} & \multicolumn{3}{c}{\small{ImageNet}}\\
 \midrule
 & MSE $\downarrow$ & PSNR $\uparrow$ & VGG f $\downarrow$ \\
 \midrule
 FF Nets & 0.0048 $\pm$ 8.89$\mathrm{e}{-5}$ & 74.5 $\pm$ 26.41 & 2.06 $\pm$ 0.56 \\ 
 SIREN  & 0.015 $\pm$ 8.4$\mathrm{e}{-4}$ & 72.1 $\pm$ 49.69 & 2.36 $\pm$ 0.73 \\ 
 \hline
 SAPE & 0.0030 $\pm$ 8.6$\mathrm{e}{-6}$ & 75.7 $\pm$ 26.26 &  \textbf{1.85$\pm$0.58} \\ 
Ours (nC) & 0.0027$\pm$1.3$\mathrm{e}{-5}$ & \textbf{76.5$\pm$24.4} & 1.92$\pm$0.59 \\ 
 Ours (eq) & 0.0028$\pm9.4\mathrm{e}{-6}$ & 75.9$\pm$24.82  & 1.91$\pm$0.54 \\
 Ours & \textbf{0.0027$\pm$1.6$\mathrm{e}{-6}$} & 75.9$\pm$23.49 & 1.89$\pm$0.49 \\ 
 \midrule
\multicolumn{1}{c}{} & \multicolumn{3}{c}{\small{DIV2K}} \\
\midrule
 & MSE $\downarrow$ & PSNR $\uparrow$ & VGG f $\downarrow$ \\
 \midrule
 FF Nets & 3.3$\mathrm{e}{-3}\pm$3.4$\mathrm{e}{-5}$ & 75.6$\pm$19.7 & 2.2$\pm$0.5   \\ 
 SIREN  & 3.1$\mathrm{e}{-3}\pm$7.6$\mathrm{e}{-6}$  & 74.5$\pm13.3$ & 2.4$\pm0.4$  \\ 
 SAPE & 2.0$\mathrm{e}{-3}\pm$2.2$\mathrm{e}{-6}$ & 76.5$\pm$13.9 & 2.1$\pm$0.5 \\ 
 Ours (nC) & 2.07$\mathrm{e}{-3}\pm$2.87$\mathrm{e}{-6}$ & 76.5$\pm$17.4 & 2.2$\pm$0.5 \\ 
 Ours (eq) & 1.91$\mathrm{e}{-3}\pm$2.46$\mathrm{e}{-6}$ & \textbf{76.9$\pm$18.9} & \textbf{2.0$\pm$0.5}  \\
 Ours & \textbf{1.88$\mathrm{e}{-3}\pm$2.26$\mathrm{e}{-6}$} & 76.8$\pm$13.6 & 2.1$\pm$0.4 \\ 
 \bottomrule
\end{tabular}}
\end{center}
\vspace{-0.5cm} 
\caption{\small{Evaluation on DIV2K and ImageNet (100 imgs)}}
\vspace{-6mm}
\label{tab:image_regression}
\end{table}
\textbf{Sample sparsity}
We qualitatively evaluate how our method reconstructs natural images for different pixel sample densities. As can be seen from Figure \ref{fig:sample_sparsity}, our model outperforms baselines for low sample densities and can achieve reasonable reconstruction accuracy even when only training on $2\%$ of the image pixels.

\textbf{Learning the base component c}
For 1D and 3D regression tasks, the scene representation is w.r.t. $0$: A 1D sinusoid oscillates around $0$ and the SDF representation of a shape consists in knowing the distance from the surface defined at $0$. For an image this is not the case, as negative RGB values have no meaning; instead, we set $c$ to the image mean. We validate this choice with following experiment: For a 2D regression task, we set $c$ to a trainable parameter in the network and optimise it together with the network weights. When initialised at different values, $c$ always converges to the image mean (see Appendix \ref{app:image_mean_experiment} for details).

\subsection{3D shape regression}
\label{sec:3D_shape_regression}
We evaluate our method on a set of 3D regression tasks. We compare against all baselines on 3D models from 3D Warehouse \cite{3DW} categories \textit{lamp, car, chair, sofa, motorbike, bed and camera}; quantitative results can be found in table \ref{tab:shape_regression}, qualitative results in Figures \ref{fig:teaser_image}, \ref{fig:3D_regression_results_motorcycle} and  \ref{fig:3D_regression_results_chair}. We observe SIREN to easily diverge on some shapes, as already noted by others \cite{Wang2021GeometryConsistentNS}. This leads to a significantly higher reconstruction error for some of the 3D shapes.  Similarly to our results in 2D, our model produces smoother reconstructions without losing detail. This is particularly visible in the example shown in Figure \ref{fig:3D_regression_results_motorcycle}.
\vspace{-1mm}
\subsection{Sampling Fourier Features at different  $\sigma$}
\label{sec:sigma_sensitivity}
\begin{figure}[ht!]
    \centering
    \begin{minipage}[c]{0.99\linewidth}
    \includegraphics[width=\linewidth]{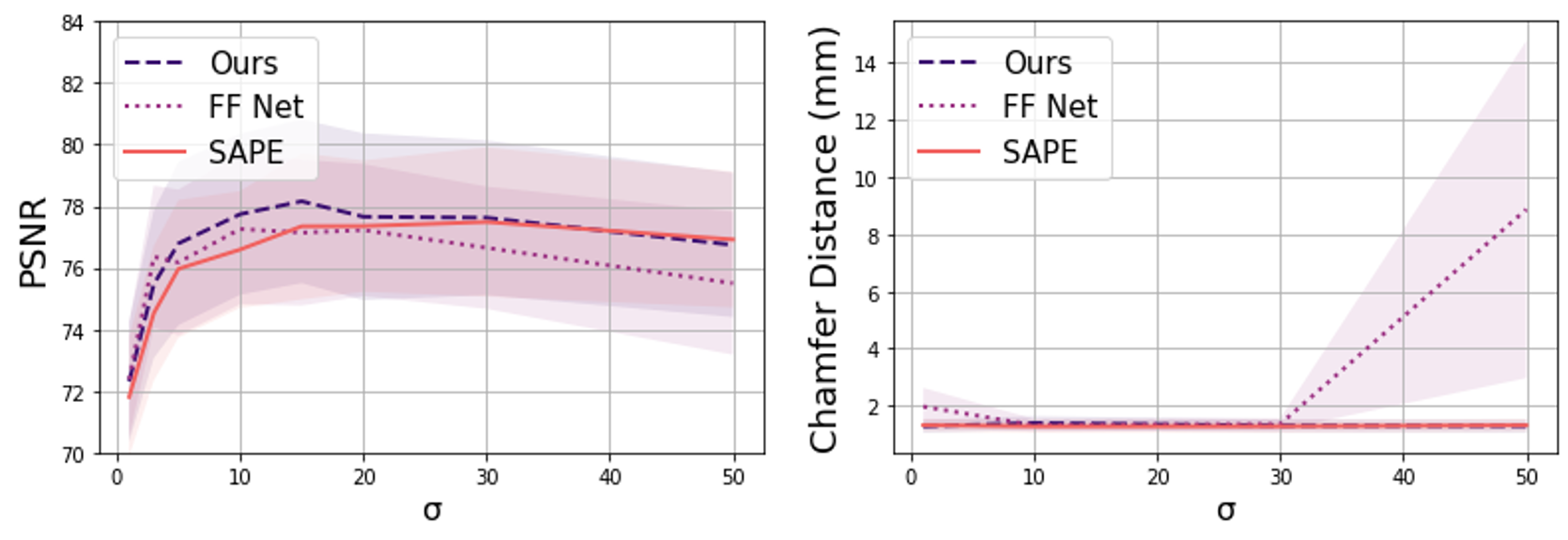}
    \end{minipage}
    \vspace{-2.5mm}
    \caption{\small{Sensitivity to $\sigma$ (standard deviation) when sampling Fourier Features for 2D regression tasks (\textbf{right}) and 3D regression tasks (\textbf{left}).}}
    \vspace{-4mm}
    \label{fig:frequency_range} 
\end{figure}

We evaluate how our model performs for different ranges of sampled frequencies on 2D and 3D regression tasks (Figure \ref{fig:frequency_range}). We evaluate for $10$ natural images from DIV2K and $10$ shapes of different categories taken from 3D Warehouse. We compare to SAPE and FF Net which also use a Fourier Feature encoding. We find that for values below $30$ our model overall outperforms both SAPE and FF Net. For values above $30$, our model is on par with SAPE without the need to mask out frequencies for smooth regions. Thanks to our progressive architecture we achieve crisper representations for low encoding frequencies and less noise is introduced in smooth regions, even when high encoding frequencies are present in the encoding (see Appendix \ref{app:qual_results_sigma} for qualitative examples). Compared to the best performance for FF Net found at $\sigma = 10$ \cite{tancik2020fourfeat}, we find that for our model, a value of $\sigma = 15$ yields the best PSNR on natural images. We attribute this to the progressive nature of our architecture that allows for higher frequencies to be used in the positional encoding without introducing noise in low structure areas. For 3D shapes we observe that for high $\sigma$ values FF Net diverges resulting in a strong drop in reconstruction accuracy. This is not the case for SAPE and our method. Although the drop in reconstruction accuracy is not significant across different values of $\sigma$, higher frequencies visibly introduce noise in the reconstruction. This can be seen in the qualitative examples presented in Appendix \ref{app:qual_results_sigma}. We find $\sigma = 3$ to be the best value for the tested 3D shapes.
\vspace{-3mm}
\vspace{-0.4mm}
\subsection{Encoding feature density}
\label{sec:frequency_density}
\begin{figure}[ht!]
    \centering
    \begin{minipage}[c]{0.49\linewidth}
    \includegraphics[width=\linewidth]{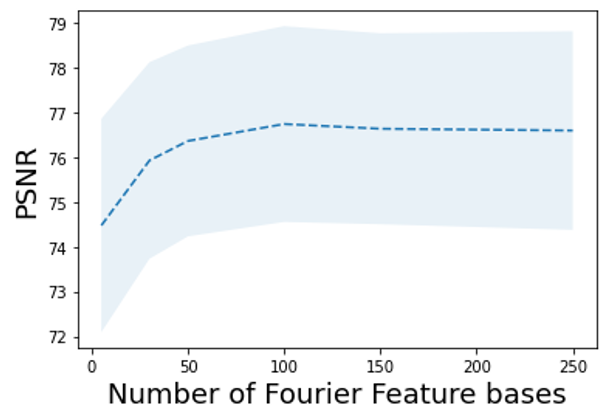}
    \end{minipage}
    \begin{minipage}[c]{0.49\linewidth}
    \includegraphics[width=\linewidth]{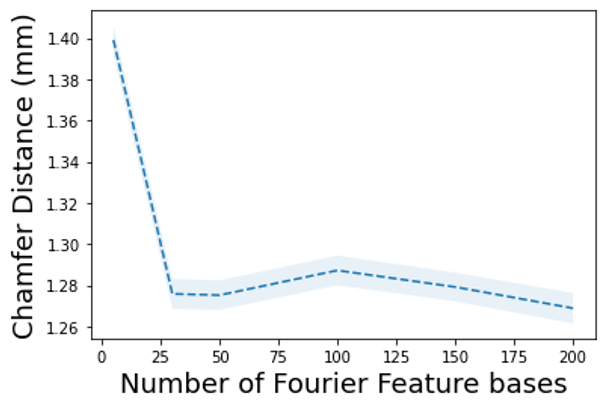}
    \end{minipage}
    \vspace{-2.5mm}
    \caption{\small{Sensitivity to frequency density of the positional encoding for 2D regression tasks (\textbf{right}) and 3D regression tasks (\textbf{left}).}}
    \vspace{-2.5mm}
    \label{fig:frequency_density} 
\end{figure}

We study how the encoding feature density (number of sampled frequencies) affects the reconstruction quality of images and 3D shapes. For a subset of 10 natural images from the DIV2K dataset and 10 shapes (of different classes) from 3D Warehouse, we plot the PSNR against the number of frequencies in the encoding (see Figure \ref{fig:frequency_density}). We set $\sigma = 15$ and $\sigma = 3$ for natural images and 3D shapes respectively. We find that the reconstruction quality saturates when using more than $100$ frequencies in the encoding for natural images and at about $30$ frequencies for 3D shapes.
\begin{figure}[ht!]
    \centering
    \begin{minipage}[c]{0.99\linewidth}
    \includegraphics[width=\linewidth]{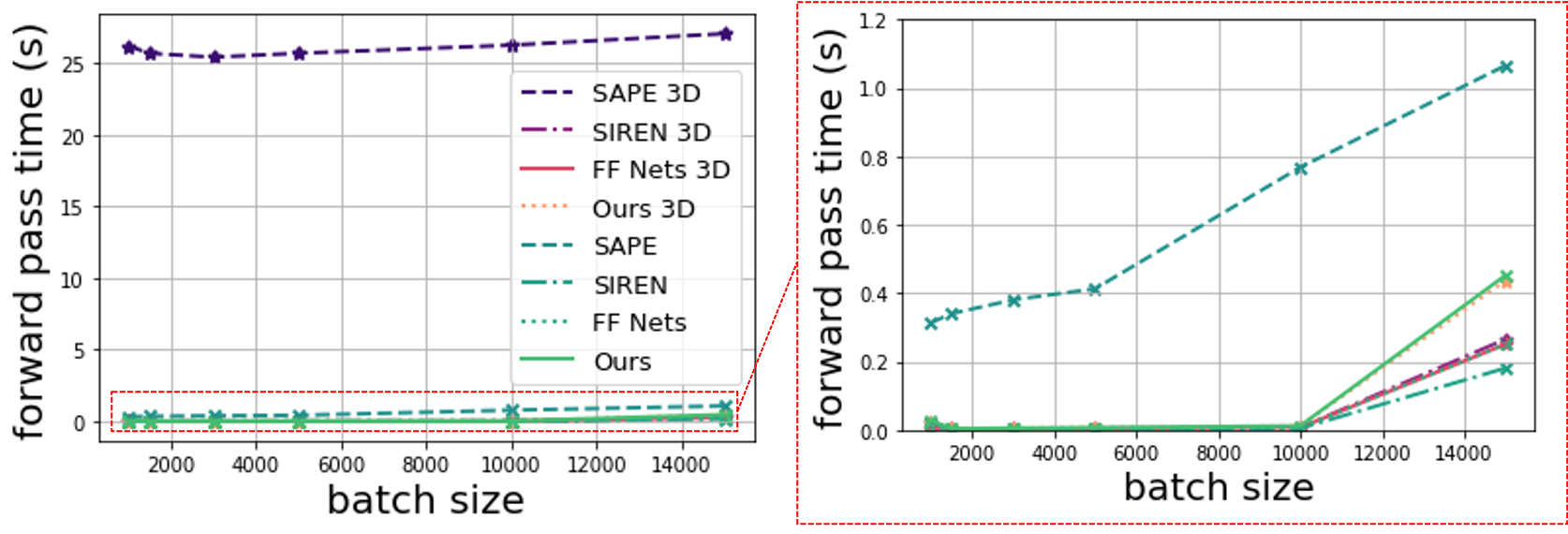}
    \end{minipage}
    \vspace{-1.5mm}
    \caption{\small{Forward pass time for 2D and 3D regression tasks at different batch sizes.}}
    \vspace{-5.5mm}
    \label{fig:training_time} 
\end{figure}

\subsection{Encoding Frequencies \& Signal Frequencies} The presented experiments as well as previous approaches \cite{tancik2020fourfeat} provide empirical results for the best values of $\sigma$ for sampling the frequencies of Fourier Features. We believe a more principled approach will select encoding frequencies based on the frequency composition of the scene itself and we provide a few initial experiments in Appendix \ref{app:FF_FT} to pave the road for future work.

\subsection{Model size and training speed}
\label{sec:model_size_and_training_time}
We evaluate our architecture in terms of its representational capacity and training speed on 2D and 3D regression tasks. We fit a subset of the DIV2K dataset for different sizes of our model (gradually increasing both network depth and width). Our method outperforms SIREN for all tested sizes and FF Net for sizes above $100k$ parameters. In particular, note how all baselines experience a performance drop as the number of parameters increases. We attribute this to overfitting due to overparameterisation. For SIREN this could also be due to instability at greater depth which we noted leads to divergence. Our model's accuracy steadily increases for the tested model sizes, suggesting that the architecture is less prone to overfitting. SAPE outperforms us in representational capacity at lower sizes, however, it is significantly slower in training speed (see Figure \ref{fig:training_time}).
SAPE's forward pass is slowed down by regular mask interpolation and update steps. This is particularly noticeable for 3D regression tasks. At the cost of a small increase in network size, our architecture can achieve the same reconstruction quality as SAPE, while training much faster and being more robust to overfitting. 
\begin{figure}[ht!]
    \centering
    \begin{minipage}[c]{0.99\linewidth}
    \includegraphics[width=\linewidth]{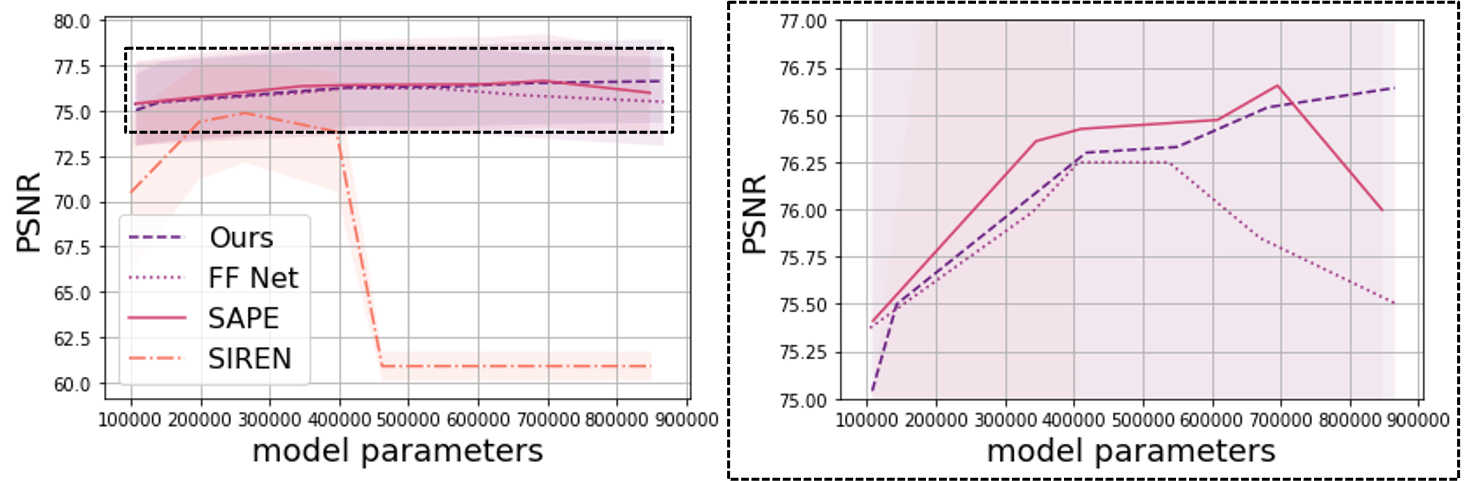}
    \end{minipage}
    \vspace{-1.5mm}
    \caption{Reconstruction accuracy wrt. model size, evaluated on a subset of the DIV2K dataset.}
    \vspace{-3mm}
    \label{fig:loss_architecture_size}
\end{figure}
\begin{figure}[ht!]
    \centering
    \begin{minipage}[c]{0.99\linewidth}
    \includegraphics[width=\linewidth]{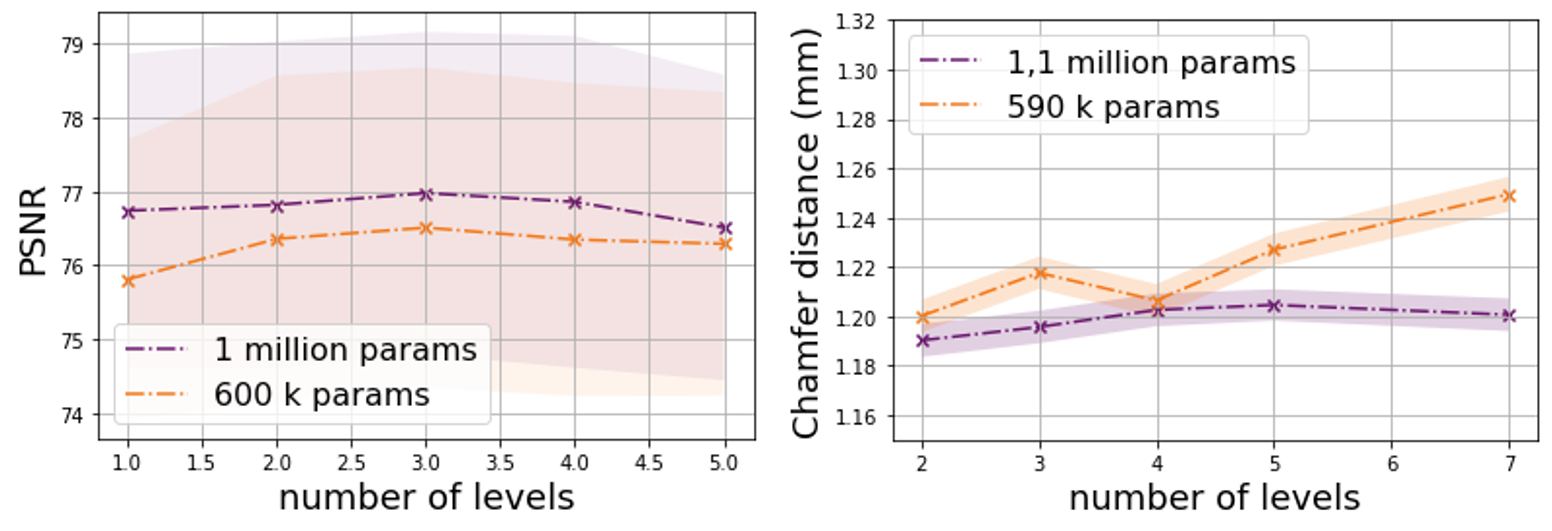}
    \end{minipage}
    \vspace{-1.5mm}
    \caption{\small{Reconstruction accuracy for different numbers of levels, keeping overall network size fixed. \textbf{Left}: 2D regression task. \textbf{Right}: 3D regression task.}}
    \vspace{-5.5mm}
    \label{fig:level_decomposition}
\end{figure}
\subsection{Ablation studies}
\subsubsection{Level composition and level conditioning}
We compare our model against an ablated version with equal weights (\textit{Ours (eq)}) in the level composition and note that while using decreasing weights leads to a better MSE score, no improvement can be seen for perceptual metrics; in fact, equal weighting leads to a slightly higher mean for PSNR and perceptual loss (see Table \ref{tab:image_regression}). However, removing the level-conditioning of higher layers on lower layers (see Equation \ref{eq:architecture}) (\textit{Ours (nC)}), leads to a more noticeable drop in performance, in particular for the high resolution images of the DIV2K dataset. 
\subsubsection{Level decomposition}
We evaluate how level decomposition affects our model's performance. Keeping the number of parameters fixed, we evaluate how the reconstruction accuracy varies with different numbers of levels. We evaluate on $10$ natural images from the DIV2K dataset and $10$ 3D models of different categories from 3D Warehouse for model sizes of roughly $1m$ and $600k$ parameters. (see Figure \ref{fig:level_decomposition}). We find that overall, within the range of levels tested, reconstruction accuracy is not strongly affected. However, we observe the following trends: for fitting natural images, $3$ levels of detail yield the best result for both tested model sizes. The lower reconstruction accuracy for more levels of detail is most likely due to the fact that each level has fewer parameters and is therefore not able to fit each scene residual well. Fewer levels of detail on the other hand may lack the representational advantage introduced by our architecture design. For fitting 3D shapes we find that for both networks, $2$ levels of detail yields the best reconstruction. The reconstruction accuracy degrades faster for increasing levels for the smaller network. We attribute this to the fact that fewer parameters are available for each level which eventually leads to underfitting.
\section{Limitations and Future Work}
We demonstrate the advantages of our proposed representation, however, several questions remain open and we would like to suggest directions for future work: 1) We select both the number of levels and the sampled frequency density manually and provide empirical studies for the best values of these hyperparameters in Sections \ref{sec:sigma_sensitivity} and \ref{sec:frequency_density}. However, it would be interesting to evaluate to what extent these parameters can be learnt as part of the representation. 2) Similarly to \cite{tancik2020fourfeat} we find the best value for $\sigma$ through parameter search, but our experiments in Appendix \ref{app:FF_FT} indicate a correlation between the frequency composition of a scene and the frequencies of the input mapping that lead to the best reconstruction. We believe that it could be beneficial to infer a scene-specific input mapping based on this correlation. 3) We did not experiment with generalisation to novel scenes within the scope of this work but believe this to be an interesting extension. 
\vspace{-2mm}
\section{Conclusion}
We introduce a novel multi-scale implicit representation based on progressive positional encoding. Through conditioning a hierarchical MLP structure on incremental Fourier Features, our method learns to decompose a scene into progressive levels of detail without level-specific supervision. We achieve higher reconstruction accuracy for 2D images and 3D shapes compared to baselines, in particular, for scenes with wide frequency spectra. The modularity of the architecture allows to only use part of the network at inference time, if a coarser representation is sufficient. Overall, our method provides a flexible, continuous and multi-scale implicit representation with a simple, end-to-end training scheme.

{\small
\bibliographystyle{ieee_fullname}
\bibliography{ms}
}

\newpage
\appendix
\onecolumn
\makeatletter
\setlength{\@fptop}{0pt}
\makeatother
\paragraph{}
\textit{We provide the following additional content: (A) complementary graphs illustrating our analysis on learning the base component $\textbf{c}$ (B) An analysis relating encoding frequencies to signal frequencies (C) Additional qualitative examples for 1D and 2D regression tasks (D) Qualitative examples for reconstructing at different values of $\sigma$. 
}

\section{Regressing the base component}
\label{app:image_mean_experiment}
In the main paper, we introduced the experiment which validates our choice for setting the base component $\textbf{c}$ of our scene composition to the image mean: we let the implicit network regress the base component as part of the optimisation. We initialise $\textbf{c}$ at different values $0.9$, $0.1$ and $0.5$ and observe that it naturally converges to the image mean (Figure \ref{fig:image mean}):
\begin{figure}[ht!]
    \centering
    \begin{minipage}[c]{0.99\linewidth}
    \includegraphics[width=\linewidth]{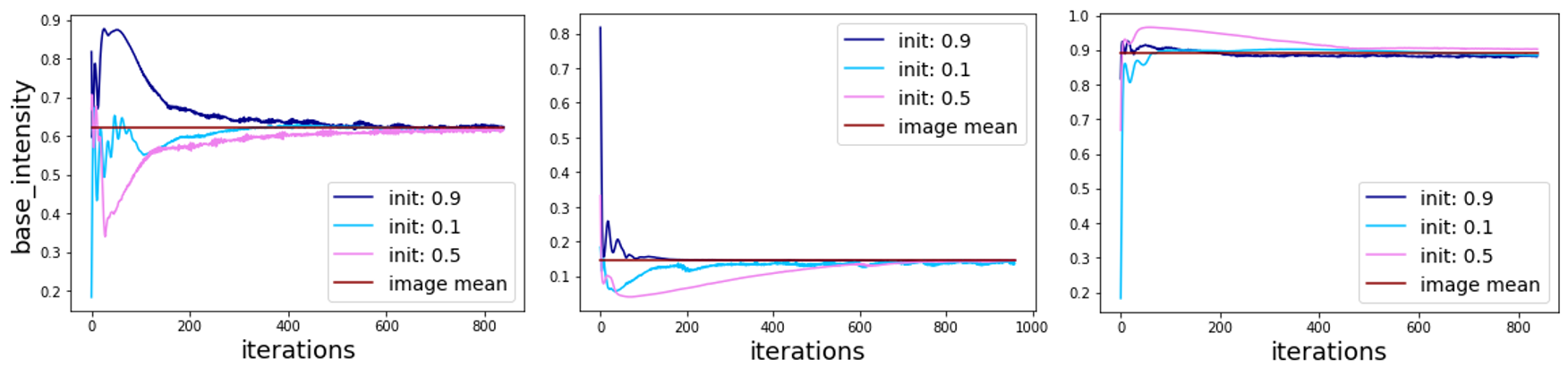}
    \end{minipage}
    \vspace{-1.5mm}
    \caption{Regressing the base intensity for 3 different images. We set the base intensity as a learnable parameter and initialise it at different values. We then optimise together with the network weights.}
    \label{fig:image mean}
    \vspace{-3mm}
\end{figure}

\section{Encoding Frequencies \& Signal Frequencies}
\label{app:FF_FT}
\begin{figure}[ht!]
    \centering
    \begin{minipage}[c]{0.99\linewidth}
    \includegraphics[width=\linewidth]{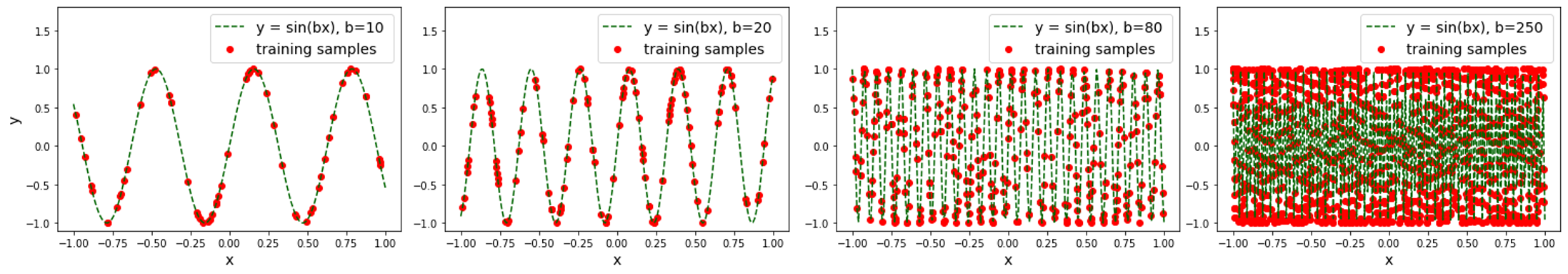}
    \end{minipage}
    \vspace{-2.5mm}
    \caption{\small{Synthetic single frequency signals used in our analysis on relating positional encoding to signal frequency (Appendix \ref{app:FF_FT})}}
    \label{fig:1D_signal_examples} 
\end{figure}

\begin{figure}[ht!]
    \centering
    \begin{minipage}[c]{0.8\linewidth}
    \includegraphics[width=\linewidth]{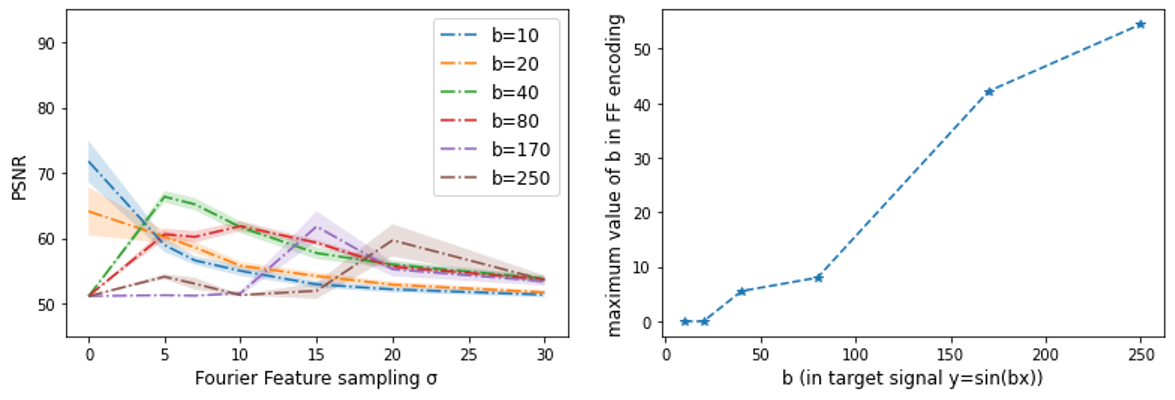}
    \end{minipage}
    \vspace{-2.5mm}
    \caption{\small{\textbf{Left}: PSNR achieved for Fourier Features (FF) sampled at different $\sigma$ for 1D synthetic signals ($y=sin(bx)$) of varying frequencies. \textbf{Right}: Highest frequency present in the FF encoding yielding the best PSNR vs. frequency of the 1D synthetic signal. }}
    \vspace{-1.0mm}
    \label{fig:FF_FT_1D} 
\end{figure}

\begin{figure}[ht!]
    \centering
    \begin{minipage}[c]{0.9\linewidth}
    \includegraphics[width=\linewidth]{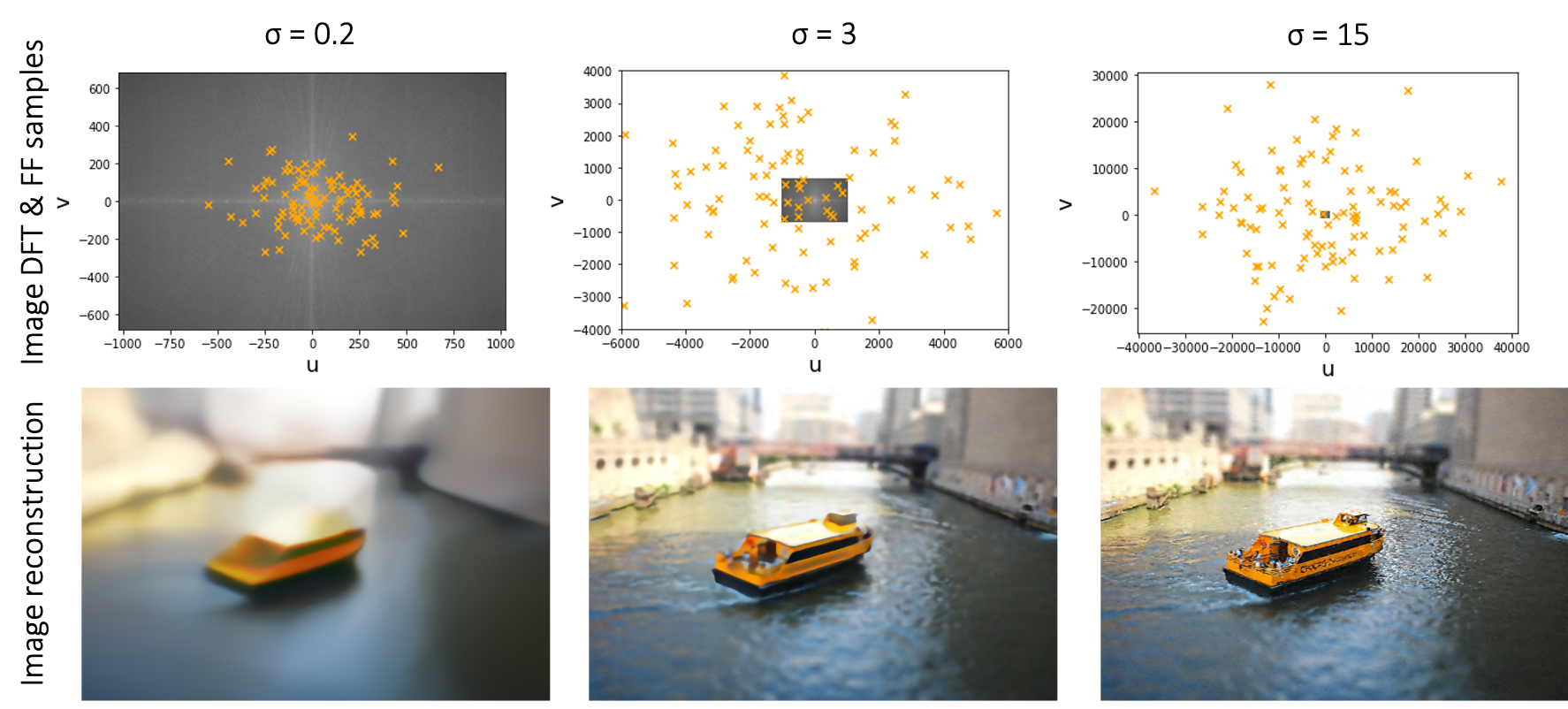}
    \end{minipage}
    \vspace{-1.5mm}
    \caption{\small{How sampled FF frequencies relate to the Discrete Fourier Transform (DFT) of a natural image. \textbf{Top}: The DFT and FF samples (orange). \textbf{Bottom}: The reconstruction .}}
    \vspace{-4mm}
    \label{fig:FF_FT_2D} 
\end{figure}

The results presented in section \ref{sec:sigma_sensitivity} demonstrate that $\sigma = 15$ is a good value to reconstruct natural images. We believe that a more principled approach to select positional encoding frequencies will leverage the frequency composition of the target signal. To pave the road for future work, we explore the relationship between the frequencies in the positional encoding and the frequencies of the signal with the following experiments: 1) For a set of synthetic 1D signals with one specific frequency (see examples in Figure \ref{fig:1D_signal_examples}), we test for which value of $\sigma$ the highest fitting accuracy is achieved. Not surprisingly, for higher frequency signals, PSNR peaks at higher $\sigma$ values (Figure \ref{fig:FF_FT_1D}, left). However, for these simple signals, the maximum frequency present inside the Fourier Features yielding the highest PSNR is overall lower than the signal frequency itself (Figure \ref{fig:FF_FT_1D}, right).
2) For a 2D regression task, we compute the DFT and visualise how the sampled Fourier basis from the Fourier Feature encoding relate to the Fourier basis functions present in the images' DFT. For a natural image, the (discrete) Fourier Transform can be written as:
\begin{equation}
    F(k,l) = \sum_{m=0}^{M-1}\sum_{n=0}^{N-1} f[m,n] e^{-j2\pi(\frac{k}{M}m+\frac{l}{N}n)}
\label{eq:2D_DFT}
\end{equation}
where $M$ and $N$ are the image dimensions.
Intuitively, the image is transformed into a set of $M \times N$ basis functions: every pixel represents one basis. Frequencies in the DFT of an image of dimension $M \times N$ range from $0$ to $\frac{m}{M}$ and $\frac{n}{N}$ \cite{2D_DFT}; the position $m,n$ of a particular Fourier Feature frequency $f_{FF}$ in the DFT domain can be computed as $m = f_{FF} M$ and $n = f_{FF} N$. For a natural image from the DIV2K dataset we plot its DFT along with the sampled Fourier Feature frequencies (Figure \ref{fig:FF_FT_2D}, top row) and display the respective reconstruction quality obtained from a simple MLP with FF positional encoding (Figure \ref{fig:FF_FT_2D}, bottom row). As the plots illustrate, when sampling frequencies from within the DFT of the image, the reconstruction is lacking high frequency detail. To obtain a detailed reconstruction, the sampled frequencies in the positional encoding need to be up to $40$ times higher than the highest pixel frequency present in the image. A complete analysis is beyond the scope of this work, but we hope to motivate future work that will establish a quantitative relationship between a signal's frequency decomposition and the positional encoding required to reconstruct it.

\begin{figure}[ht!]
    \centering
    \begin{minipage}[c]{0.99\linewidth}
    \includegraphics[width=\linewidth]{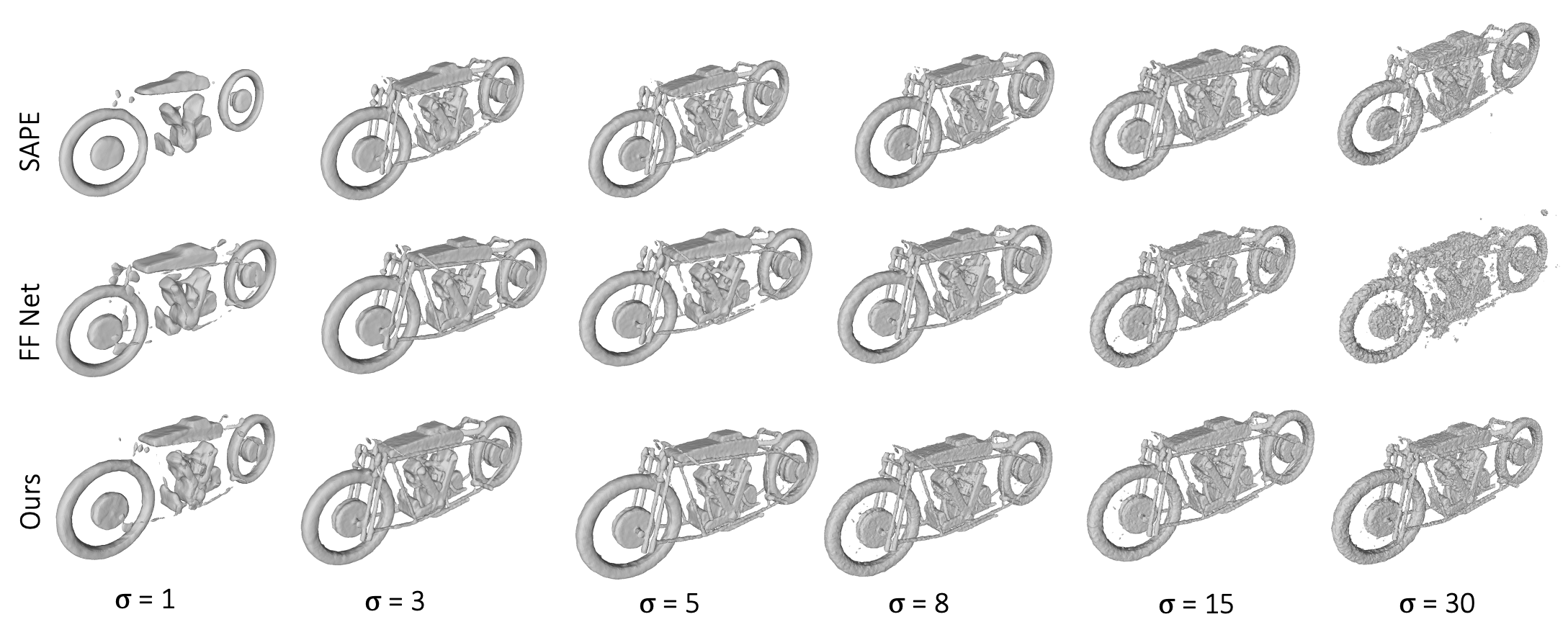}
    \end{minipage}
    \vspace{-1.5mm}
    \caption{\small{Fitting a 3D shape (3D Warehouse) with Fourier Features sampled at different values of $\sigma$.}}
    \label{fig:sigma_3D_exp}
\end{figure}
\section{Additional Qualitative Results (1D and 2D regression)}
\label{app:1D_2D_qual_results}
\begin{figure}[ht!]
    \centering
    \begin{minipage}[c]{0.99\linewidth}
    \includegraphics[width=\linewidth]{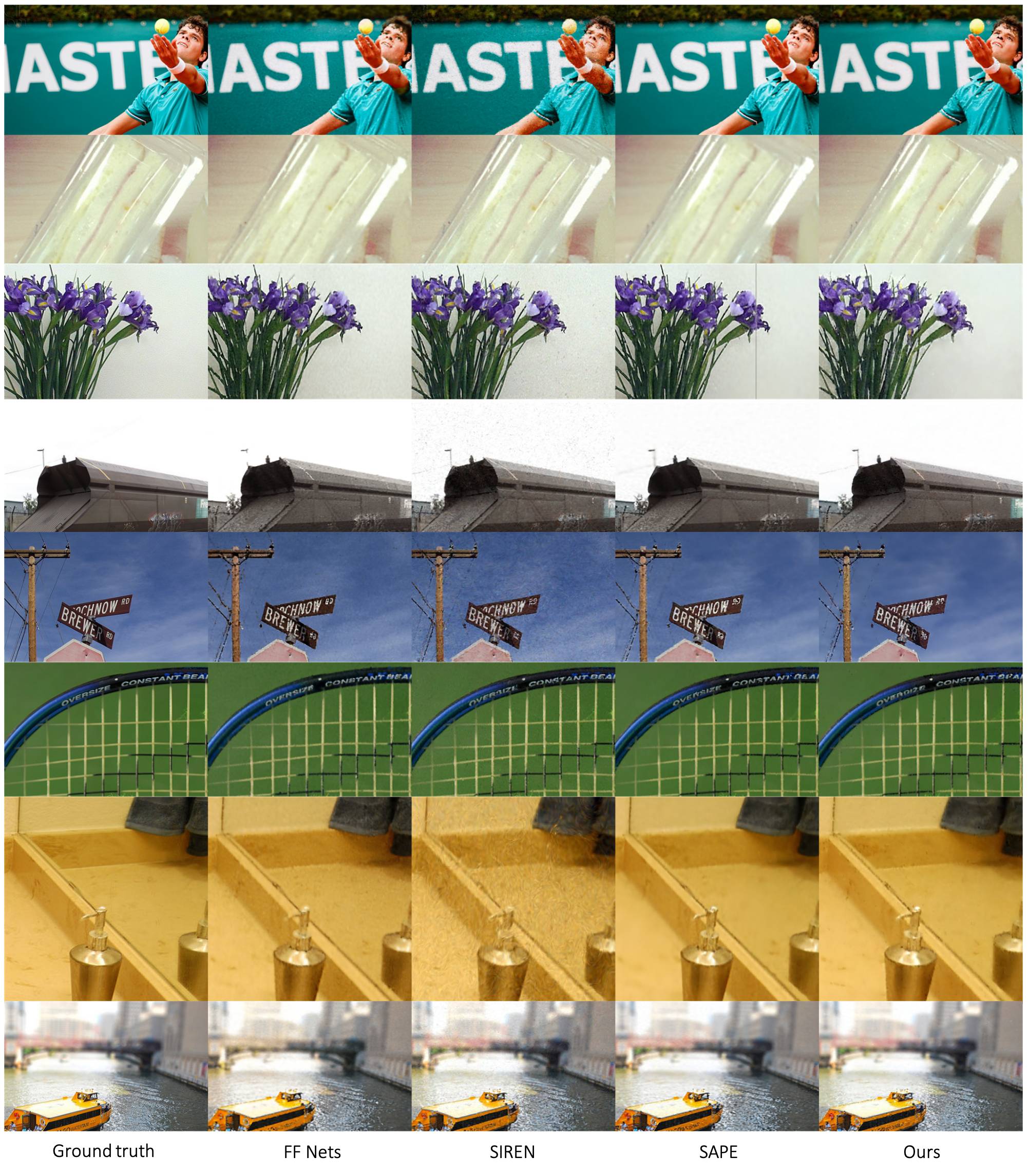}
    \end{minipage}
    \vspace{-2.5mm}
    \caption{\small{Qualitative results on the $2017$ COCO validation set.}}
    \label{fig:qual_2D_comparison}
\end{figure}

\begin{figure}[ht]
    \centering
    \begin{minipage}[c]{0.99\linewidth}
    \includegraphics[width=\linewidth]{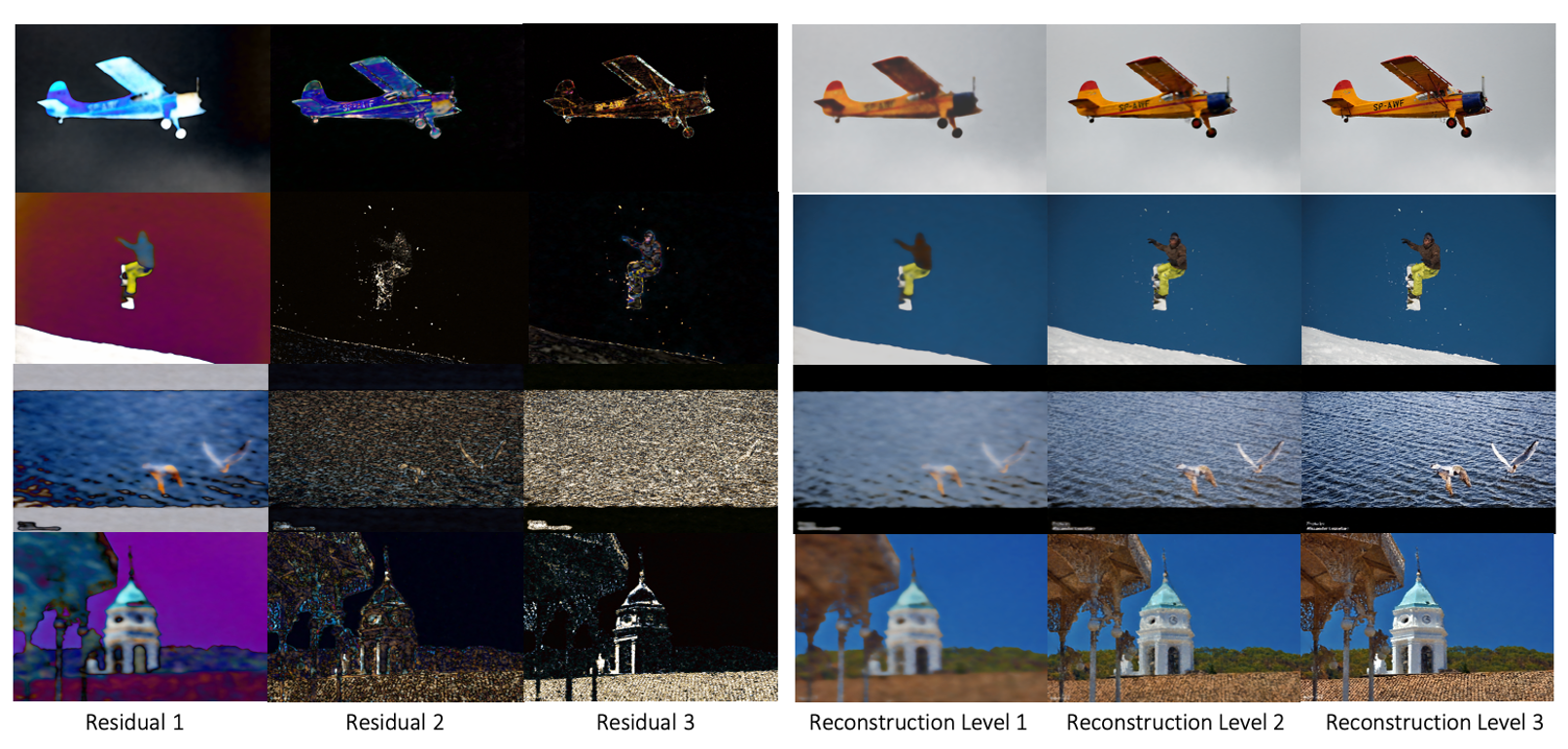}
    \end{minipage}
    \vspace{-2.5mm}
    \caption{\small{Additional qualitative examples showing our models residual output as well as the incremental level reconstructions. Images are taken from the COCO $2017$ validation dataset \cite{lin2015microsoft}.}}
    \label{fig:qual_2D_residuals}
\end{figure}

\begin{figure}[ht!]
    \centering
    \begin{minipage}[c]{0.99\linewidth}
    \includegraphics[width=\linewidth]{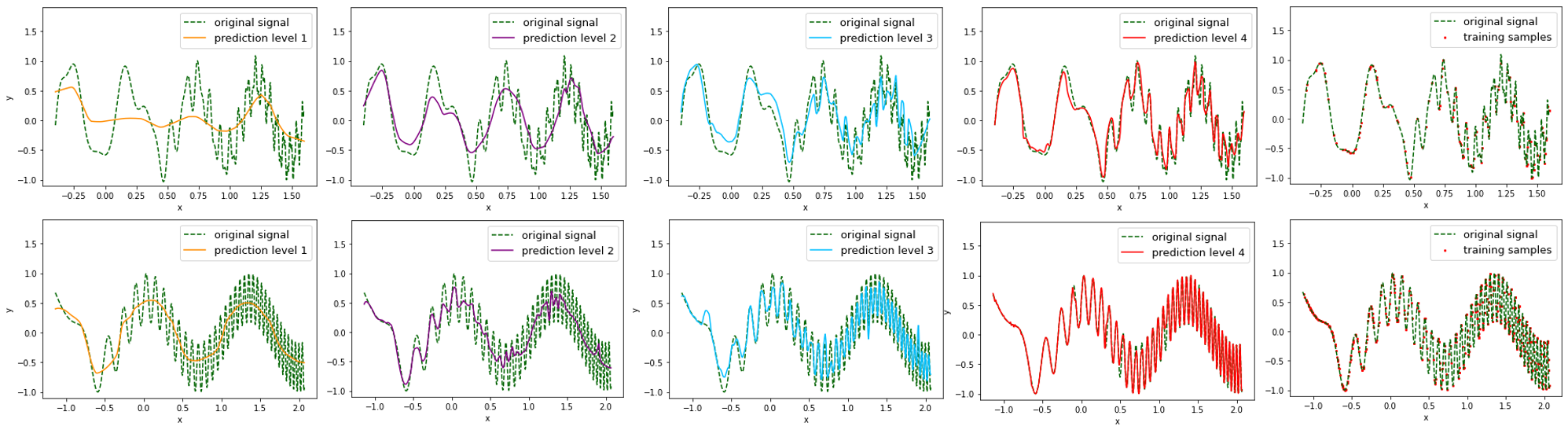}
    \end{minipage}
    \vspace{-1.5mm}
    \caption{\small{Regressing 1D periodic signals composed of multiple sinusoids.\textbf{Left} to \textbf{right}: Reconstruction Levels 1-4 and original signal with training samples.}}
    \label{fig:Additional_1D_exp}
\end{figure}

In Figure \ref{fig:Additional_1D_exp} we present additional qualitative results for regressing 1D periodic signals and Figure \ref{fig:qual_2D_comparison} shows additional qualitative results on the $2017$ COCO validation set for our method and baselines. In Figure \ref{fig:qual_2D_residuals} we display more examples of the predicted residuals $R_{1 \dots 3}$ by the network and the reconstructions $S_{1 \dots 3}$ at progressive level of detail. 
\section{Qualitative Results for reconstructing with Fourier Features sampled at different $\sigma$}
\label{app:qual_results_sigma}
\begin{figure}[ht!]
    \centering
    \begin{minipage}[c]{0.99\linewidth}
    \includegraphics[width=\linewidth]{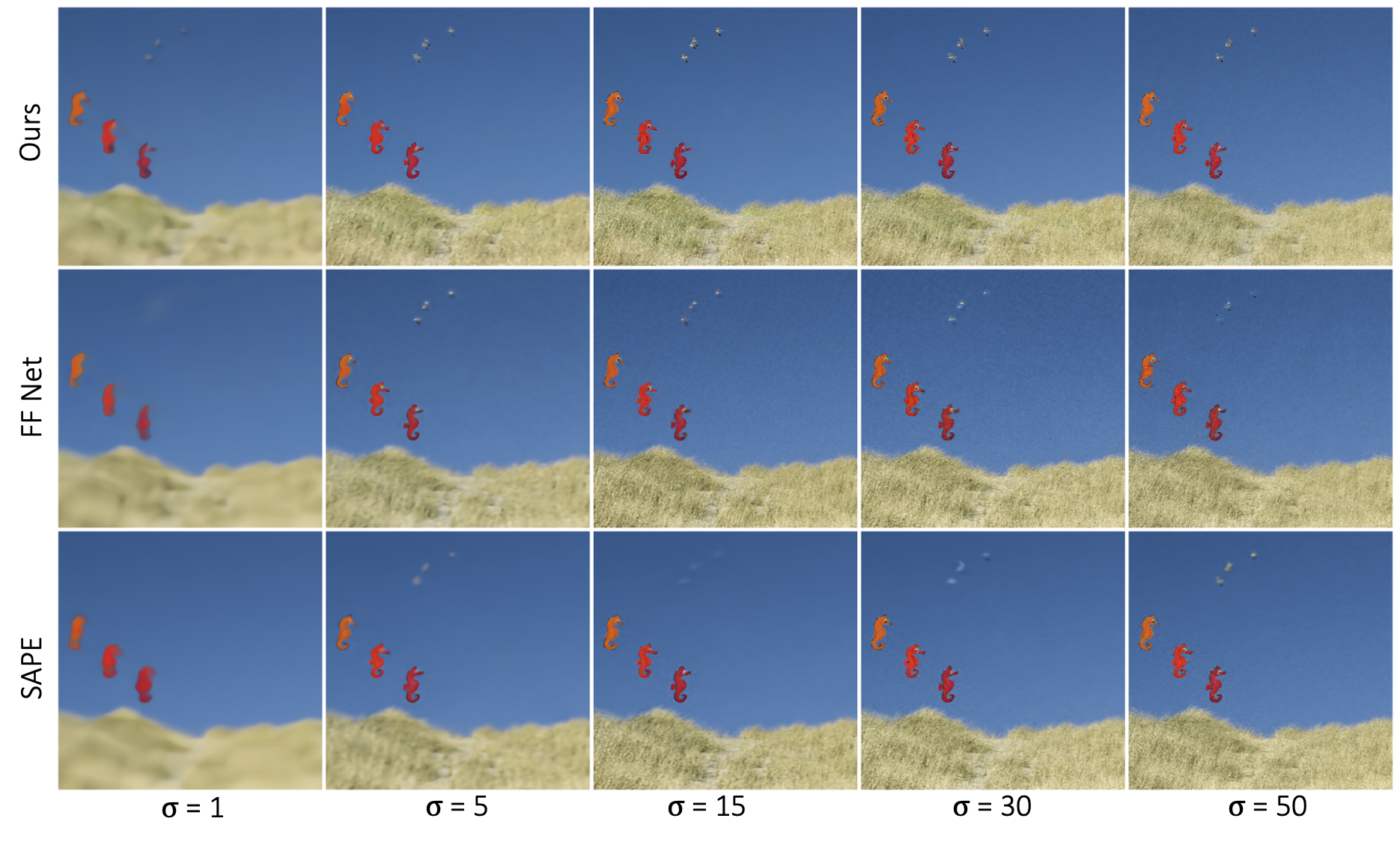}
    \end{minipage}
    \vspace{-3.5mm}
    \caption{\small{Regressing a 2D image ($2017$ COCO validation set) with Fourier Features sampled at different values of $sigma$}}
    \label{fig:qual_2D_sigma}
\end{figure}

We provide qualitative results for reconstructing 2D and 3D scenes with Fourier Features sampled at different values of $\sigma$. Note how our model provides overall crisper natural image reconstructions over a wide range of $\sigma$ (Figure \ref{fig:qual_2D_sigma}). For 3D shapes (Figure \ref{fig:sigma_3D_exp}), our model has a reconstruction quality similar to that of SAPE, even for encodings with very high frequencies (sampled at $\sigma > 15$). Compared to FF Net, the reconstruction does not diverge for high encoding frequencies (see middle row at $\sigma = 30$). The reconstructions for high $\sigma$ values remain noisy however and we experimentally find that the best reconstructions are obtained for $\sigma$ values between $3$ and $5$.

\end{document}